\documentclass[letterpaper, 10 pt, conference]{ieeeconf}  
\usepackage{amssymb}   
\usepackage{amsmath}
\usepackage{graphicx}
\usepackage{booktabs}
\usepackage[table,xcdraw]{xcolor}
\definecolor{customblue}{HTML}{EEEEEE} 
\usepackage{multirow}
\usepackage{makecell}

\usepackage{hyperref}
\usepackage{algorithm}
\usepackage{algorithmicx}
\usepackage{algpseudocode}

\IEEEoverridecommandlockouts                              

\title{\LARGE \bf
DECAMP: Towards Scene-Consistent Multi-Agent Motion Prediction with Disentangled Context-Aware Pre-Training 
}
\author{Author Names Omitted for Anonymous Review}

\author{Jianxin Shi$^{1}$, Zengqi Peng$^{2}$, Xiaolong Chen$^{2}$, Tianyu Wo$^{1}$, Jun Ma$^{2,3}$, \textit{Senior Member, IEEE}
\thanks{$^{1}$Jianxin Shi and Tianyu Wo are with the School of Computer Science and Engineering, Beihang University, Beijing, China
}%
\thanks{$^{2}$ Zengqi Peng, Xiaolong Chen, and Jun Ma are with Robotics and Autonomous Systems Thrust, The Hong Kong University of Science and Technology (GZ), Guangzhou, China
}%
\thanks{$^{3}$ Jun Ma is with Division of Emerging Interdisciplinary Areas, The Hong Kong University of Science and Technology, Hong Kong, China
}%
}

\begin{document}

\maketitle
\thispagestyle{empty}
\pagestyle{empty}


\begin{abstract} Trajectory prediction is a critical component of autonomous driving, essential for ensuring both safety and efficiency on the road. However, traditional approaches often struggle with the scarcity of labeled data and exhibit suboptimal performance in multi-agent prediction scenarios. To address these challenges, we introduce a disentangled context-aware pre-training framework for multi-agent motion prediction, named DECAMP. Unlike existing methods that entangle representation learning with pretext tasks, our framework decouples behavior pattern learning from latent feature reconstruction, prioritizing interpretable dynamics and thereby enhancing scene representation for downstream prediction. Additionally, our framework incorporates context-aware representation learning alongside collaborative spatial-motion pretext tasks, which enables joint optimization of structural and intentional reasoning while capturing the underlying dynamic intentions. Our experiments on the Argoverse 2 benchmark showcase the superior performance of our method, and the results attained underscore its effectiveness in multi-agent motion forecasting. To the best of our knowledge, this is the first context autoencoder framework for multi-agent motion forecasting in autonomous driving. The code and models will be made publicly available. \end{abstract}

\section{Introduction}

Since accurately predicting the movements of traffic agents is crucial for safe planning, motion prediction has witnessed substantial advances in recent years in the context of autonomous driving~\cite{huang2022survey,chang2019argoverse,wilson2023argoverse,bharilya2024machine}.
However, this task remains highly challenging due to the inherent complexity of individual behavioral patterns and the mutual influence of their dynamic interactions~\cite{karle2022scenario}. 
Consequently, inadequate modeling of multi-agent joint motion can generate trajectory combinations for target agents that are inconsistent with the overall scene, ultimately hindering the ego vehicle from making reliable decisions. 
\begin{figure}[thbp]
    \centering
\includegraphics[width=0.85\linewidth]{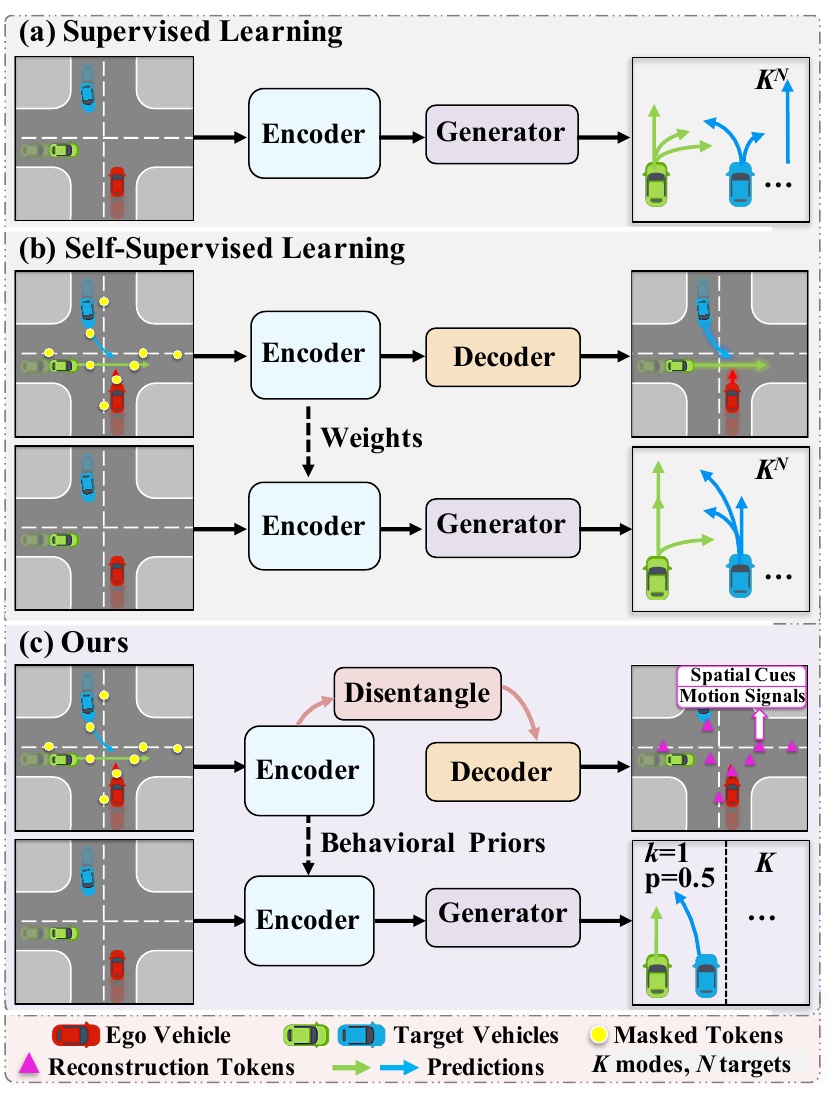}
    \caption{Key difference between previous methods and ours. (a) Supervised learning methods require costly labeled data and provide limited encoder representational capacity. (b) Existing self-supervised learning methods focus on single-agent prediction and tightly couple the encoder with pretext tasks. (c) Our method supports joint prediction and disentangles encoder representation learning from decoder pretext task execution. }
    \label{fig:intro}
    \vspace{-5mm}
\end{figure} 

The mainstream solutions of trajectory prediction can be broadly categorized into supervised learning and self-supervised learning methods. 
The primary pipeline of supervised learning-based approaches is shown in Fig.~\ref{fig:intro}(a), where graph neural networks~\cite{rowe2023fjmp} or transformer-based architectures~\cite{zhou2022hivt} serve as encoders to process input scenes. 
These models effectively capture complex agent interactions and integrate map semantics~\cite{wilson2023argoverse} to generate a unified scene representation, which is subsequently fed into the generator. 
The generator has evolved from anchor-free designs~\cite{wang2023spatio} that directly regress future coordinates to more complex anchor-based methods, such as goal-conditioned~\cite{xin2025multi} and heatmap-based~\cite{gilles2022gohome} approaches. 
These advancements substantially enhance prediction accuracy and behavioral plausibility. 
However, the reliance on supervised training remains a limitation due to the high cost of collecting labeled data. 
Therefore, recent research has shifted toward self-supervised trajectory prediction. 
The core principle is to enhance encoder representations via mask-reconstruction pretext tasks, followed by transferring the pre-trained encoder to downstream prediction, thereby improving overall performance, as shown in Fig.~\ref{fig:intro}(b). 

However, self-supervised trajectory prediction methods still encounter two core challenges. 
First, the ``\textit{encoder-decoder}" pre-training paradigm results in strong entanglement between the behavioral features learned by the encoder during pre-training and the features required for the reconstruction task. This coupling prevents the encoder from focusing on driving behavior representations, thereby limiting improvements in downstream trajectory prediction accuracy. 
Second, these methods are typically designed for single-agent prediction. When extended to $N$-agent prediction, each target agent requires separate scene rotation and translation operations, followed by individual prediction. 
The resulting $N$ sets of $K$ predicted trajectories are then combined, requiring a complex post-processing procedure to filter the $K^{N}$ trajectory combinations and ensure that the final top-$K$ sets preserve scene-level consistency. Clearly, this paradigm is impractical for multi-agent joint prediction. 

To address these challenges, we propose a \textbf{D}is\textbf{E}ntangled \textbf{C}ontext-\textbf{A}ware pre-training framework for \textbf{M}ulti-agent \textbf{P}rediction (DECAMP), which follows an ``\textit{encoder–regressor–decoder}" cascade paradigm as shown in Fig.~\ref{fig:intro}(c). 
Unlike previous methods that entangle representation learning with auxiliary pretext tasks, our approach emphasizes disentangled, behavior-centric pre-training by adapting the regressor~\cite{chen2024context} from masked image modeling to behavior prediction in autonomous driving. 
By integrating both spatial cues and motion signals, the model acquires richer behavioral priors, thereby enhancing its ability to interpret latent traffic dynamics and generalize more effectively to downstream multi-agent forecasting. 
The main contributions of this work are summarized as follows: 
\begin{itemize}
\item We introduce a disentangled self-supervised learning framework that guides the encoder toward robust prior knowledge, thereby enhancing scene representation for downstream prediction tasks. 
\item We design collaborative spatial–motion pretext tasks that jointly optimize spatial cue reconstruction and motion signal recognition, enabling the model to capture positional structures and reveal underlying dynamic intentions. 
\item Extensive experiments on the Argoverse 2 motion forecasting dataset~\cite{wilson2023argoverse} demonstrate that our method significantly improves predictive accuracy, particularly in challenging tasks involving complex multi-agent interactions. 
\end{itemize}

\section{Related Work} 
\subsection{Supervised Learning for Motion Forecasting} 
The effective representation of scene elements is crucial for accurately predicting the behavior of agents. 
Early approaches rasterize map information and agent states into images~\cite{casas2018intentnet}, extracting spatio-temporal features via convolutional networks~\cite{lecun1998convolutional}. 
However, rasterization often overlooks critical topological information, such as lane connectivity and spatial dependencies among agents, which limits the predictive accuracy of models. 
Consequently, methods like DenseTNT~\cite{gu2021densetnt} and HiVT~\cite{zhou2022hivt} introduce vectorized modeling for single-agent prediction. 
Notably, these methods focus on supervised trajectory prediction, generating vectorized representations of maps and agents, incorporating additional attributes into these vectors, and leveraging graph neural networks or transformers for scene encoding. 
Such techniques enable a comprehensive preservation of topological structures and interactions. 
Building on this progress, we adopt similar vectorized representations for joint prediction. 

\subsection{Self-Supervised Learning for Motion Forecasting} 
Inspired by the success of self-supervised learning in computer vision, recent studies have extended this paradigm to single-agent behavior prediction in autonomous driving. These approaches are broadly classified into generative and discriminative methods. Representative generative methods include RMP~\cite{yang2023rmp}, Forecast-MAE~\cite{cheng2023forecast}, Forecast-PEFT~\cite{wang2024forecast} and Social-MAE~\cite{11128614}. Their core idea is to design diverse masking strategies and apply them to trajectory sequences and map data during pre-training, to reconstruct the positions of masked tokens. 
Discriminative methods such as DTO~\cite{monti2022many}, PreTraM~\cite{xu2022pretram}, and Behavior-Pred~\cite{shi2025behavior} apply varying degrees of data augmentation, including masking or synthesis during pre-training. 
They subsequently employ intra-modal or cross-modal contrastive learning to capture latent interactions, thereby obtaining a unified representation of traffic elements within the scene. 

However, both methods typically entangle representation learning with the optimization of pretext tasks during pre-training, resulting in impure behavioral priors. Consequently, directly transferring such encoders to downstream fine-tuning tasks may constrain performance. 
Prior research in masked image modeling has shown that a well-designed regressor~\cite{chen2024context} can partially mitigate this entanglement in image pre-training. 
Motivated by this insight, we extend this approach to develop a novel pre-training framework for multi-agent behavior prediction and evaluate its effectiveness in autonomous driving scenarios. 
Unlike previous methods, our study first reveals that collaborative pretext tasks, which integrate both spatial cues and motion signals, can enhance the representation learning of driving behavior patterns.

\section{Method} 
\begin{figure*}[thbp]
    \centering
\includegraphics[width=0.96\textwidth]{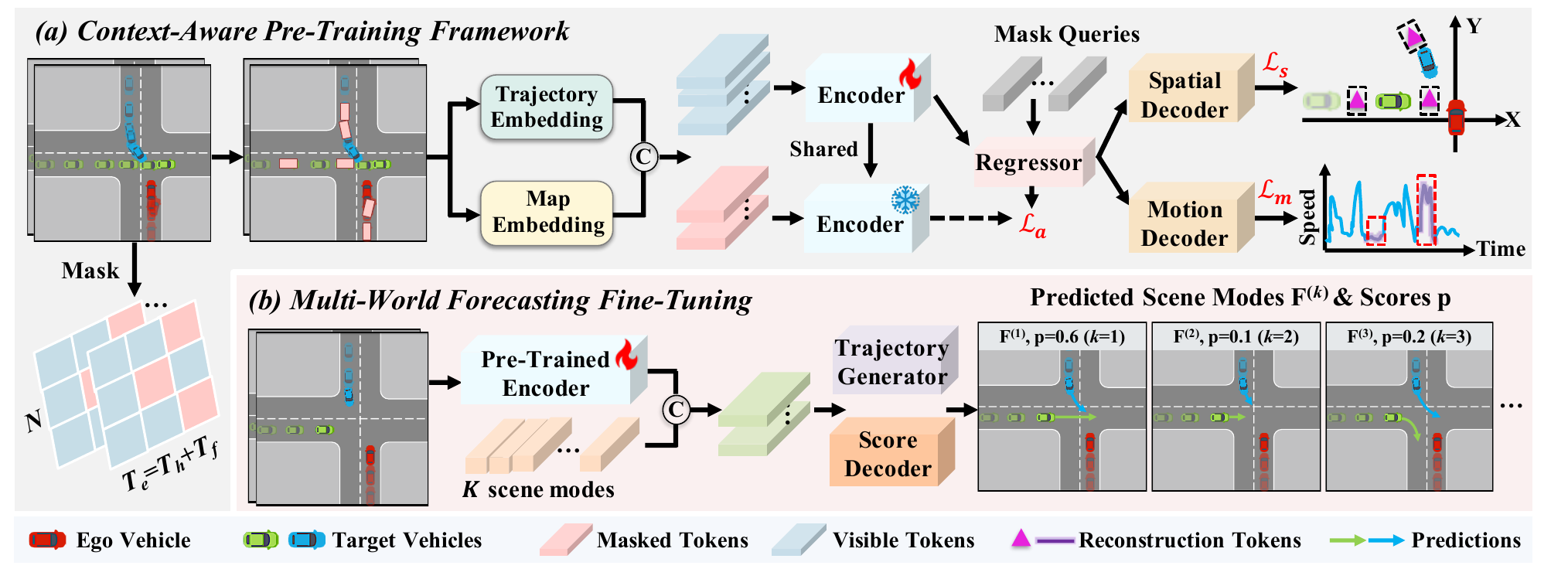}
    \caption{Overview of DECAMP: (a) The pre-training stage takes historical and future states of traffic agents, and map elements as inputs. Through pretext tasks that combine spatial reconstruction with motion recognition, the encoder is guided to learn disentangled representations of behavior, while the decoder focuses on completing the pretext objectives. (b) The fine-tuning stage uses only historical states and map elements as inputs. By leveraging the robust behavioral priors learned during pre-training, the model generates scene-consistent joint predictions for downstream multi-agent trajectory prediction.}
    \label{fig:architecture}
\end{figure*} 
We propose DECAMP, a pre-training approach for motion prediction which can prevent the encoder from being entangled with specific pretext tasks, as illustrated in Fig.~\ref{fig:architecture}. 
To effectively capture driving behavior patterns, we introduce two specialized pretext tasks. These tasks individually decode spatial cues and motion signals within the scene context, thereby guiding the optimization of encoder representation learning. Section~\ref{subsec1} defines the multi-agent motion forecasting task and outlines its inputs and outputs. 
Section~\ref{subsec2} provides a modular description of our proposed DECAMP and details its optimization objective. 
Finally, Section~\ref{subsec3} explains how fine-tuning leverages behavioral priors for joint prediction with scene-level consistency. 
\subsection{Problem Formulation} 
\label{subsec1} 
Multi-agent motion forecasting aims to predict the joint future motions of multiple interacting agents by leveraging their historical states $\mathbf{S}$ and the map context $\mathbf{M}$. 
To reduce the sensitivity of the prediction model to variations in the reference coordinate system, we rotate and translate the entire scene with the ego vehicle as the coordinate origin. This process can be formalized as Algorithm~\ref{alg:coord-transform}. All subsequent definitions and representations are established under this unified coordinate system. 
\begin{algorithm}[t]
\caption{Coordinate Transformation for Agent $i$}
\label{alg:coord-transform}
\begin{algorithmic}[1]
\Require Agent position $\widetilde{\mathbf{a}}_i=(\widetilde{x}_i, \widetilde{y}_i)$ and heading $\widetilde{\psi}_i$,
      ego current position $\mathbf{o}=(x_0, y_0)$ and heading $\theta$ 
\Ensure Transformed position $\mathbf{a}_i$ and heading $\psi_i$ 

\State {Translation:} Compute relative position $\mathbf{a}_i' \gets \widetilde{\mathbf{a}}_i - \mathbf{o}.$

\State {Rotation:} Construct rotation matrix 
\[
\mathbf{I}(\theta) = 
\begin{bmatrix}
\cos(\theta) & -\sin(\theta) \\
\sin(\theta) & \cos(\theta)
\end{bmatrix}.
\]

\State {Transform position:}      $\mathbf{a}_i \gets \mathbf{a}_i' \mathbf{I}(\theta).$

\State {Transform heading:} $\psi_i \gets \mathrm{wrap}\!\left(\widetilde{\psi}_i-\theta\right)$,
\Statex where $\mathrm{wrap}(\cdot)$ normalizes the angle into $(-\pi,\pi]$.

\State \Return $\mathbf{a}_i$, $\psi_i$
\end{algorithmic}
\end{algorithm} 

The observed historical states are represented as $\mathbf{S}=\{(x_{n}^{t},y_{n}^{t})\mid n=1,2,\dots,N\}_{t=1}^{T_{h}} $, where $N$ is the number of agents, $(x_{n}^{t},y_{n}^{t})$ denotes the position of $n$-th agent at time $t$, and $T_{h}$ is the observation horizon. 
The map context is defined as $\mathbf{M}=\{L_{z} \mid z=1,2, \dots, Z\}$, which contains $Z$ lane segments. 
Each lane segment $L_{z}=\{s_{w} \mid w=1,2,\dots, W\}$ represents a polyline with $W$ centerline waypoints. 
Based on these inputs, the objective is to predict $K$ plausible future scenarios, each represented by a set of joint trajectories for all agents: $\mathbf{F}=\{\mathcal{F}^{(k)}\mid k=1,2,\dots, K\}$, $\mathcal{F}^{(k)}=\{(\hat{x} _{n}^{t},\hat{y} _{n}^{t})\mid n=1,2,\dots,N\}_{t=T_{h}+1}^{u}$, where $u=T_{h}+T_{f}$, $T_{f}$ is the prediction horizon. 
The ground truth is $\mathbf{Y}=\{(x_{n}^{t},y_{n}^{t})\mid n=1,2,\dots,N\}_{t=T_{h}+1}^{u}$. 

\subsection{Context-Aware Pre-Training Framework} 
\label{subsec2}
Our self-supervised pre-training framework DECAMP follows an ``\textit{encoder–regressor–decoder}" pipeline, as illustrated in Fig.~\ref{fig:architecture}(a). 
The encoder extracts static environmental features and models interactions among traffic elements, generating context-aware scene representations. 
A regressor is introduced between the encoder and decoder to predict latent representations of masked tokens from visible token information. This design enhances the ability of the model to capture semantic relationships among scene elements and provides the encoder with effective representation optimization objectives. 
Finally, the dual-decoder utilizes the predicted representations of masked tokens to perform two pretext tasks: spatial reconstruction and motion recognition. 

\textbf{Input.} 
The input of the model consists of the observed trajectories $\mathbf{S}\in\mathbb{R}^{N\times T_{h} \times D}$ of agents, their corresponding future trajectories $\mathbf{Y}\in\mathbb{R}^{N\times T_{f} \times D}$, and relevant map elements $\mathbf{M}\in\mathbb{R}^{Z\times W \times D}$. 
To facilitate fine-grained representation learning, a temporal random masking strategy is applied to both agent trajectories and lane segments. 
The resulting visible and masked tokens are then projected into $D$-dimensional feature space through modality-specific embedding layers. 
Specifically, we employ a Feature Pyramid Network (FPN)~\cite{lin2017feature} for trajectories and PointNet~\cite{qi2017pointnet} for map elements: 
\begin{equation}
\mathbf{S}_{v},\mathbf{Y}_{v},\mathbf{M}_{v}=\mathrm{Mask}(\mathbf{S},\mathbf{Y},\mathbf{M}), 
\label{eq1}
\end{equation} 
\begin{equation}
\bar{\mathbf{S}}_{v},\bar{\mathbf{Y}}_{v}=\mathrm{FPN}(\mathbf{S}_{v},\mathbf{Y}_{v}), \bar{\mathbf{S}}_{v},\bar{\mathbf{Y}}_{v}\in\mathbb{R}^{N\times D}, 
\label{eq2}
\end{equation} 
\begin{equation}
\bar{\mathbf{M}}_{v}=\mathrm{PointNet}(\mathbf{M}_{v}), \bar{\mathbf{M}}_{v}\in\mathbb{R}^{Z\times D}, 
\label{eq3}
\end{equation} 
where $\mathbf{S}_{v}\in\mathbb{R}^{N\times(1-r_{1})T_{h} \times D}$, 
$\mathbf{Y}_{v}\in\mathbb{R}^{N\times(1-r_{2})T_{f} \times D}$, and $\mathbf{M}_{v}\in\mathbb{R}^{Z\times(1-r_{3})W \times D}$ denote visible tokens, the parameters $r_{1}$, $r_{2}$ and $r_{3}$ indicate the corresponding mask ratios. 
For simplicity, representations of masked tokens $\bar{\mathbf{S}}_{m}$, $\bar{\mathbf{Y}}_{m}$ and $\bar{\mathbf{M}}_{m}$ are computed in the same manner. 

\textbf{Encoder.} 
After obtaining the embedding representations of the visible and mask tokens, they are independently processed by the siamese encoder $\mathcal{P}_{e}$, which is implemented using a stack of transformer encoder layers. 
Each encoder integrates trajectory and map information to generate a context-aware representation of the scene. 
These representations effectively capture the behavioral patterns of agents and model the dynamic interactions among different traffic elements. 
The context representations for visible and masked tokens are formally defined as: 
\begin{equation}
\mathbf{E}_{v}=\mathcal{P}_{e}(\mathrm{Concat}(\bar{\mathbf{S}}_{v},\bar{\mathbf{Y}}_{v},\bar{\mathbf{M}}_{v}) + \mathrm{PE}), 
\label{eq5}
\end{equation} 
\begin{equation}
\mathbf{E}_{m}=\mathcal{P}_{e}(\mathrm{Concat}(\bar{\mathbf{S}}_{m},\bar{\mathbf{Y}}_{m},\bar{\mathbf{M}}_{m}) + \mathrm{PE}), 
\label{eq6}
\end{equation} 
where $\mathbf{E}_{v},\mathbf{E}_{m}\in\mathbb{R}^{(N+N+Z) \times D}$, $\mathrm{PE}\in\mathbb{R}^{D}$ is the position encoding. 

\textbf{Regressor.} 
The context regressor comprises a stack of Multi-Head Attention (MHA) modules, designed to perform regression prediction of masked token representations based on visible context during self-supervised pre-training. 
Specifically, each cross-attention module employs learnable mask queries $\mathbf{Q}_{m}\in\mathbb{R}^{(N+N+Z) \times D}$ as the query, while the visible token representations $\mathbf{E}_{v}$ serve as the key and value. This process generates latent regression representations $\mathbf{R}_{m}$ for the masked tokens: 
\begin{equation}
\mathbf{R}_{m}=\mathrm{MHA}(\mathrm{Q}=\mathbf{Q}_{m},\mathrm{K}=\mathbf{E}_{v},\mathrm{V}=\mathbf{E}_{v}). 
\label{eq7}
\end{equation} 
By leveraging contextual information from visible tokens, the model effectively reconstructs representations of masked tokens, thereby enhancing its semantic modeling capacity. 
Meanwhile, an alignment loss is introduced to ensure the encoder focuses on learning robust representations of driving behavior patterns, while the decoder specializes in accomplishing the pre-training objective: 
\begin{equation}
\mathcal{L}_\mathrm{a} = \ell_{r}(\mathbf{R}_{m}, \mathbf{E}_{m}),  
\label{eq8}
\end{equation} 
where we use MSE loss for $\ell_{r}(\mathbf{R}_{m}, \mathbf{E}_{m})$.

\textbf{Decoder.} 
The decoder converts the latent regression representation $\mathbf{R}_{m}$ of the masked tokens into the target representation for the pretext task, thereby avoiding direct reliance on information contained in the visible tokens. 
To enable a more comprehensive analysis of traffic behavior, a dual-decoder architecture is introduced to perform two pretext tasks: the spatial decoder $\mathcal{G}_{s}$ that reconstructs the spatial cues of masked tokens, and the motion decoder $\mathcal{G}_{m}$ that identifies their motion states. 
Similar to the encoder, each decoder comprises a stack of transformer blocks based on self-attention mechanisms. 
The decoders share a common backbone for the spatial and motion branches, while two distinct linear prediction heads are used to generate their respective outputs: 
\begin{equation}
\hat{\mathbf{S}}_\mathrm{pos},\hat{\mathbf{Y}}_\mathrm{pos},\hat{\mathbf{M}}=\mathrm{Linear}(\mathcal{G}_{s}(\mathbf{R}_{m})), 
\label{eq82}
\end{equation} 
\begin{equation}
\hat{\mathbf{S}}_\mathrm{mot},\hat{\mathbf{Y}}_\mathrm{mot}=\mathrm{Linear}(\mathcal{G}_{m}(\mathbf{R}_{m})), 
\label{eq9}
\end{equation} 
where $\hat{\mathbf{S}}_\mathrm{pos},\hat{\mathbf{Y}}_\mathrm{pos}\in\mathbb{R}^{N \times D}$, $\hat{\mathbf{M}}\in\mathbb{R}^{Z \times D}$ represent the predicted spatial cue representations for the masked tokens, while $\hat{\mathbf{S}}_\mathrm{mot}\in\mathbb{R}^{N \times (r_{1}T_{h})}$ and $\hat{\mathbf{Y}}_\mathrm{mot}\in\mathbb{R}^{N \times (r_{2}T_{f})}$ correspond to the predicted motion signals. 
Decoding the representation $\mathbf{R}_{m}$ of the regressor imposes an implicit constraint on the encoder. This process guides the encoder to learn behaviorally relevant features that are crucial for motion prediction and well aligned with downstream tasks. 

\textbf{Pre-Training Objective.} 
For the spatial decoder $\mathcal{G}_{s}$ and motion decoder $\mathcal{G}_{m}$, we define reconstruction losses $\mathcal{L}_{s}$ and $\mathcal{L}_{m}$ to support the model in solving pretext tasks: 
\begin{equation}
\mathcal{L}_{s} = \ell_{s}(\hat{\mathbf{S}}_\mathrm{pos}, \bar{\mathbf{S}}_{m})+\ell_{y}(\hat{\mathbf{Y}}_\mathrm{pos}, \bar{\mathbf{Y}}_{m})+\ell_{m}(\hat{\mathbf{M}}, \bar{\mathbf{M}}_{m}),
\label{eq10}
\end{equation} 
\begin{equation}
\mathcal{L}_{m} = \ell_{sv}(\hat{\mathbf{S}}_\mathrm{mot}, \bar{\mathbf{S}}_\mathrm{mot})+\ell_{yv}(\hat{\mathbf{Y}}_\mathrm{mot}, \bar{\mathbf{Y}}_\mathrm{mot}),   
\label{eq11}
\end{equation} 
where $\bar{\mathbf{S}}_\mathrm{mot}\in\mathbb{R}^{N \times (r_{1}T_{h})}$ and $\bar{\mathbf{Y}}_\mathrm{mot}\in\mathbb{R}^{N \times (r_{2}T_{f})}$ denote the instantaneous speed of the masked tokens, which provides crucial kinematic information for trajectory prediction. 
And all reconstruction terms are implemented using the L1 loss. 
The overall pre-training objective combines these terms with the alignment loss as: 
\begin{equation}
\mathcal{L}_\mathrm{pre\text{-}train} = \alpha \mathcal{L}_\mathrm{a}+\mathcal{L}_{s}+ \mathcal{L}_{m}, 
\label{eq12}
\end{equation} 
where $\alpha$ serves as a weighting coefficient that balances the relative contributions of the alignment objective and two pretext tasks. 

\subsection{Multi-World Forecasting Fine-Tuning} 
\label{subsec3} 
Our fine-tuning stage for multi-agent prediction adopts an ``\textit{encoder–generator}" pipeline, as illustrated in Fig.~\ref{fig:architecture}(b). 
A key distinction from pre-training is that the model receives only historical states and map information, without access to future states. This setup aligns with real-world application requirements. 
Furthermore, instead of predicting independent trajectories per agent, our model generates $K$ distinct scene hypotheses. 
Each hypothesis corresponds to a complete set of joint trajectories for all target agents, ensuring scene-level consistency by maintaining mutual motion compatibility and avoiding unrealistic behaviors such as collisions. 

\textbf{Encoder.} 
Through self-supervised pre-training, the encoder $\mathcal{P}_{e}$ learns robust representations of driving behavior patterns. 
For downstream fine-tuning on the multi-agent motion forecasting task, the encoder is initialized by transferring weights from the pre-trained $\mathcal{P}_{e}$ to incorporate behavioral priors: 
\begin{equation}
\mathbf{Z}_{e}= \mathcal{P}_{e}(\mathrm{Concat}(\mathrm{FPN}(\mathbf{X}),\mathrm{PointNet}(\mathbf{M} )) +\mathrm{PE} ),  
\label{eq13}
\end{equation} 
where $\mathbf{Z}_{e}\in\mathbb{R}^{(N+Z) \times D}$ denotes the contextual representation of the current scene. 

\begin{figure*}[!tb]
    \centering
\includegraphics[width=0.86\textwidth]{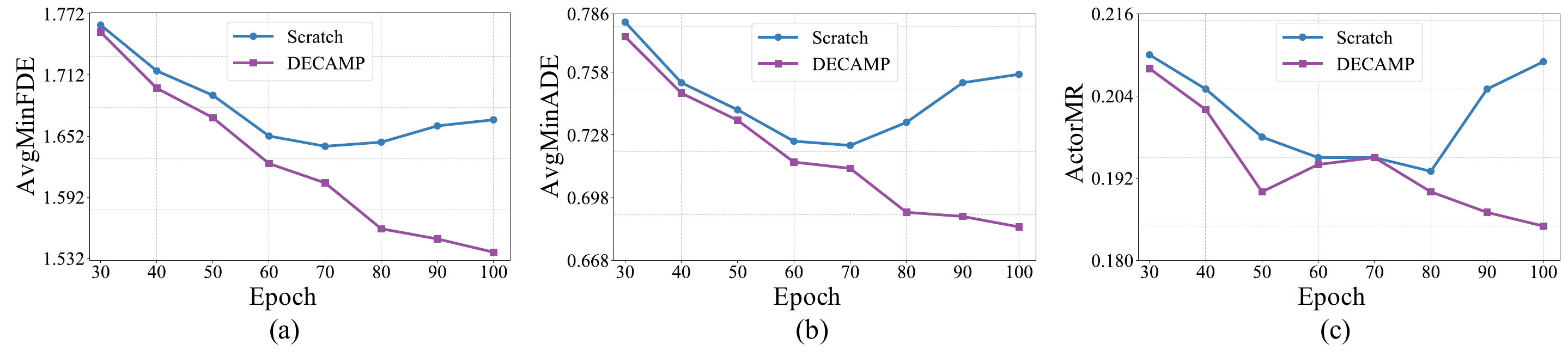}
    \caption{Comparison of our proposed DECAMP with the model trained from scratch at different training epochs.}
    \label{fig:results}
    \vspace{-0.5em}
\end{figure*} 

\textbf{Generator.} 
To further evaluate the effectiveness of the pre-trained encoder, the scene representation $\mathbf{Z}_{e}$ is employed to predict $K$ distinct scene modes along with their corresponding probability scores through a lightweight anchor-free trajectory generator and a score decoder. 
Specifically, the trajectory generator $\mathcal{G}_{t}$ is implemented using a multilayer perceptron (MLP), and the score decoder $\mathcal{G}_{p}$ comprises a linear prediction head. 
The prediction process is formulated as: 
\begin{equation}
\mathbf{F}=\mathcal{G}_{t}(\mathrm{Concat}(\mathbf{Z}_{e},\mathbf{m}_{s})), 
\label{eq14}
\end{equation} 
\begin{equation}
\mathbf{p}=\mathcal{G}_{p}(\mathrm{Concat}((\mathbf{Z}_{e},\mathbf{m}_{s})), 
\label{eq15}
\end{equation} 
where $\mathbf{m}_{s}\in\mathbb{R}^{K \times D}$ denotes a set of learnable mode-specific embeddings. 
$\mathbf{F}\in\mathbb{R}^{K \times N\times T_{f}\times 2}$ and $\mathbf{p}\in\mathbb{R}^{K}$ represent the predicted scene modes and their corresponding scores, respectively. Each scene mode includes a set of trajectory combinations for $N$ agents. 

\textbf{Fine-Tuning Objective.} 
During supervised fine-tuning, the loss function is composed of both regression and classification terms. 
To simultaneously improve trajectory prediction accuracy and mitigate mode collapse, we adopt a winner-takes-all strategy for joint optimization. 
In particular, the optimal mode is determined by selecting the candidate that minimizes the final displacement error between the multi-modal predictions $\mathbf{F}$ and the ground truth $\mathbf{Y}$: 
\begin{equation}
k^* = \arg\min_{k} \frac{1}{N}\sum_{n=1}^{N} \left\Vert \mathbf{F}^{(k)}_{n,u} - \mathbf{Y}_{n,u} \right\Vert_2,  
\label{eq16}
\end{equation} 
where $k\in\{1,2,\dots,K\}$, and $u=T_{h}+T_{f}$ denotes the final prediction timestep. 
After determining the optimal mode $k^*$, we use Huber loss for regression and Cross-Entropy (CE) loss for classification: 
\begin{equation}
\mathcal{L}_\mathrm{fine\text{-}tune}=\mathrm{Huber}(\mathbf{F}^{(k^*)},\mathbf{Y})+\mathrm{CE}(k^*,\mathbf{p}). 
\label{eq19}
\end{equation}

\section{Experiments} 
\subsection{Experimental Setup}
\textbf{Dataset.} 
Our experiments use the Argoverse 2 motion forecasting dataset~\cite{wilson2023argoverse}, which comprises 250,000 driving scenarios collected across six geographically diverse cities and uniformly sampled at 10 Hz. 
Each scenario lasts 11 seconds, consisting of 5 seconds of observation ($T_h$=50) followed by 6 seconds of prediction ($T_f$=60). 
Argoverse 2 extends its predecessor~\cite{chang2019argoverse} with more diverse scenarios and a comprehensive taxonomy of 10 distinct object classes covering both static and dynamic agents. 

\textbf{Evaluation Metrics.} 
We evaluate our method using three standard metrics for multi-agent motion forecasting: Average Minimum Final Displacement Error (AvgMinFDE), Average Minimum Displacement Error (AvgMinADE), and actor Miss Rate (ActorMR). 

The AvgMinFDE denotes the mean Final Displacement Error (FDE) of a predicted world, averaged across all target agents within a scenario: 
\begin{equation}
\mathrm{AvgMinFDE}=\min_{k} \frac{1}{N}\sum_{n=1}^{N}\left\Vert \mathbf{F}^{(k)}_{n,u} - \mathbf{Y}_{n,u} \right\Vert_2. 
\label{eq20}
\end{equation} 

The AvgMinADE represents the mean Average Displacement Error (ADE) of a predicted world, again summarized across all target agents in the scenario: 
\begin{equation}
\mathrm{AvgMinADE}=\min_{k} \frac{1}{N}\frac{1}{T_{f}}\sum_{n=1}^{N} \sum_{t=T_{h}+1}^{u}\left\Vert \mathbf{F}^{(k)}_{n,t} - \mathbf{Y}_{n,t} \right\Vert_2. 
\label{eq21}
\end{equation} 

The ActorMR is defined as the proportion of agent predictions in the $k^*$ predicted world whose FDE exceeds the threshold $\tau$: 
\begin{equation}
\mathrm{ActorMR} = \frac{1}{N} \sum_{n=1}^{N} \mathbb{I}\left\{ 
\left\Vert \mathbf{F}^{(k^*)}_{n,u} - \mathbf{Y}_{n,u} \right\Vert_2 > \tau 
\right\},  
\label{eq22}
\end{equation} 
where $\mathbb{I}\{\cdot\}$ is the indicator function, and $\tau=2.0$. 

\textbf{Implementation Details.} 
Our proposed DECAMP is trained on two NVIDIA A100 GPUs. The model undergoes 100 epochs of pre-training. Training uses synchronous batch normalization with a batch size of 32, a learning rate of $3\times10^{-3}$, and a weight decay of $1\times10^{-2}$. 
For multi-agent prediction, a dropout rate of 0.1 is adopted across all transformer blocks. 

\subsection{Quantitative Analysis} 
\textbf{Main Results.} 
\begin{table}[t]
\caption{Comparison with multi-agent methods on Argoverse 2 validation set (upper group) and leaderboard (lower group).}
\label{mainresults_multi}
\vspace{-1.7em}
\begin{center}
\begin{tabular}{cccc}
\toprule
\multirow{2}{*}[-0.7ex]{\makecell{Methods}} & \multicolumn{3}{c}{Multi-World Performance ($K$=6)} 
\\
\cmidrule(lr){2-4} & AvgMinFDE $\downarrow$   & AvgMinADE $\downarrow$   & ActorMR    $\downarrow$  \\
\midrule
MultiPath~\cite{chai2019multipath}    & 2.13          & 0.89          & 0.33          \\
FJMP (Non-F)~\cite{rowe2023fjmp} & 1.96          & 0.83          & 0.34          \\
FJMP~\cite{rowe2023fjmp}         & 1.92          & 0.82          & 0.33          \\
Forecast-MAE~\cite{cheng2023forecast} & 1.64          & 0.72          & 0.19          \\
\rowcolor{customblue}
DECAMP        & \textbf{1.53}    & \textbf{0.68}    & \textbf{0.18}    \\
 \bottomrule
 \toprule
\multirow{2}{*}[-0.7ex]{\makecell{Methods}} & \multicolumn{3}{c}{Multi-World Performance ($K$=6)}
\\
\cmidrule(lr){2-4}     & AvgMinFDE  $\downarrow$  & AvgMinADE $\downarrow$   & ActorMR   $\downarrow$   \\
\midrule
LaneGCN~\cite{liang2020learning}      & 3.24          & 1.49          & 0.37          \\
HiVT~\cite{zhou2022hivt}         & 2.20          & 0.88          & 0.26          \\
FJMP~\cite{rowe2023fjmp}         & 1.89          & 0.81          & 0.23          \\
FFINet~\cite{kang2024ffinet}       & 1.77          &  0.77          & 0.24          \\
 Forecast-MAE~\cite{cheng2023forecast}      &  1.68        & 0.73         &0.20          \\
MIND~\cite{li2024multi}      &   1.62       &  0.70        &   0.20       \\
\rowcolor{customblue}
DECAMP         & \textbf{1.57}    & \textbf{0.69}    & \textbf{0.19} \\ 
 \bottomrule
\end{tabular}
\end{center}
\vspace{-4mm}
\end{table} 
We first evaluate the performance of DECAMP on the Argoverse 2 multi-world prediction benchmark. As shown in Table~\ref{mainresults_multi}, DECAMP consistently outperforms other multi-agent prediction methods across three metrics. 
Specifically, compared with the self-supervised method Forecast-MAE~\cite{cheng2023forecast}, 
which adopts an ``\textit{encoder-decoder}" pre-training paradigm, 
DECAMP reduces AvgMinFDE by 6.70\%, AvgMinADE by 5.56\%, and ActorMR by 5.26\% on the validation set. 
Notably, lower values indicate smaller discrepancies between $K$ predicted scene modes and real-world scenarios, thereby reflecting superior performance. In the official leaderboard evaluation, these metrics decrease by 6.55\%, 5.47\%, and 5.00\%, respectively. These results demonstrate the effectiveness of DECAMP’s disentangled pre-training paradigm combined with spatial–motion collaborative pretext tasks in addressing complex multi-agent joint prediction. 

Since we adopt an ego-centric global reference coordinate system to normalize the scene during preprocessing, while the prediction task focuses on the target agent, it might be expected that our performance would be inferior to methods using the agent-centric system. 
To verify this assumption, we further evaluate DECAMP on the Argoverse 2 single-agent prediction benchmark, as shown in Table~\ref{mainresults_single}. 
For consistency, we use three official metrics: MinFDE, MinADE, and MR. 
The results show that DECAMP achieves competitive performance even under this setting. In particular, compared with Forecast-MAE~\cite{cheng2023forecast}, which employs an agent-centric reference system, DECAMP reduces MinFDE by 2.14\% and MinADE by 1.41\%. These findings further confirm the effectiveness of DECAMP in both multi-world and single-agent prediction. 
\begin{table}[!tb]
\caption{Comparison with single-agent methods on Argoverse 2 validation set (upper group) and leaderboard (lower group).}
\label{mainresults_single}
\vspace{-0.8em}
\begin{center}
\begin{tabular}{cccc}
\toprule
\multirow{2}{*}[-0.7ex]{\makecell{Methods}} & 
\multicolumn{3}{c}{Single-Agent Performance ($K$=6)} 
\\
\cmidrule(lr){2-4}& MinFDE $\downarrow$   & MinADE $\downarrow$   & MR    $\downarrow$ \\
\midrule
 MultiPath~\cite{chai2019multipath}   &  2.13 &  0.89        &  0.33         \\
 DenseTNT~\cite{gu2021densetnt}&      1.71     &  1.00         &    0.22     \\
  Argo2goalmp~\cite{gomez2023improving}      &    1.42      &    0.81    &  -         \\
Forecast-MAE~\cite{cheng2023forecast} &    1.40       &  0.71        &   0.18       \\
GANet~\cite{wang2023ganet} &     1.40      &    0.81      &   -       \\
\rowcolor{customblue}
DECAMP        & \textbf{1.37}    & \textbf{0.70}    & \textbf{0.17}    \\
 \bottomrule
\toprule
\multirow{2}{*}[-0.7ex]{\makecell{Methods}} & \multicolumn{3}{c}{Single-Agent Performance ($K$=6)}
\\
\cmidrule(lr){2-4}& MinFDE $\downarrow$   & MinADE $\downarrow$   & MR    $\downarrow$  \\
\midrule
  GISR~\cite{wei2025goal}    &   2.28        &  1.03      &    0.34      \\
  LaneGCN~\cite{liang2020learning}      &   1.96       & 0.91       &    0.30       \\
 FRM~\cite{park2023leveraging}       &   1.81        &    0.89       &  0.29        \\
  DenseTNT~\cite{gu2021densetnt}     &   1.66        &0.99            &   0.23        \\
    HDGT~\cite{jia2023hdgt}     &    1.60       &0.84            &   0.21        \\
        THOMAS~\cite{gilles2021thomas}     &   1.51        &   0.88         &   0.20        \\
\rowcolor{customblue}
DECAMP         & \textbf{1.44}    & \textbf{0.73}    & \textbf{0.18} \\ 
 \bottomrule
\end{tabular}
\end{center}
\end{table} 
\begin{table}[!tb]
\centering
\caption{Ablation study of our proposed components.}
\label{ablation_Components}
\begin{tabular}{cccccccccc}
\toprule
\multicolumn{3}{c}{Components} & \multicolumn{3}{c}{Multi-World Performance ($K$=6)}       \\
\midrule
CRA        & SCR        & MSR        & AvgMinFDE & AvgMinADE & ActorMR  \\
\midrule
         &          &          &  1.668 &0.757  & 0.209  
    \\
              \checkmark    &     \checkmark    &         &  1.592    &  0.708      &  0.189       \\
       \checkmark  &          &   \checkmark       &  1.551    & 0.687 &   0.187    \\
               &      \checkmark     &   \checkmark       &  1.603    &0.703  & 0.191      \\
    \rowcolor{customblue}
        \checkmark         &   \checkmark     & \checkmark         &  \textbf{1.534}  & \textbf{0.684} &  \textbf{0.183}
         \\
         \bottomrule
\end{tabular}
\vspace{-2mm}
\end{table} 

\textbf{Comparison with Scratch-Trained Baseline.} 
Fig.~\ref{fig:results} compares our fine-tuned model based on DECAMP with the scratch-trained baseline. As training progresses, the performance gap between the two models widens, reaching its peak at epoch 100. 
Specifically, our method reduces AvgMinFDE by 8.03\%, AvgMinADE by 9.64\%, and ActorMR by 12.44\% compared to the scratch-trained model. 
These improvements demonstrate that DECAMP effectively captures behavioral representations during pre-training and supplies valuable driving priors for the downstream prediction task. 
\begin{table}[!tb]
\centering
\caption{Ablation results of different encoder depths.}
\label{ablation_encoder}
\begin{tabular}{cccccccccc}
\toprule
\multirow{2}{*}[-0.5ex]{\makecell{Encoder \\ Depth}}
& \multicolumn{3}{c}{Multi-World Performance ($K$=6)}       \\
\cmidrule(lr){2-4}
 &   AvgMinFDE & AvgMinADE & ActorMR  \\
\midrule
     2     &   1.552  & 0.689 & 0.187  
    \\
       \rowcolor{customblue}
  \textbf{4}   &    \textbf{1.534}     &  \textbf{0.684}    & \textbf{0.183}
         \\
   6      &  1.538 &\textbf{0.684} &  0.185
   \\
         \bottomrule
\end{tabular}
\end{table}
\begin{table}[!tb]
\centering
\caption{Ablation results of different regressor depths.}
\label{ablation_regressor}
\begin{tabular}{cccccccccc}
\toprule
\multirow{2}{*}[-0.5ex]{\makecell{Regressor \\ Depth}}
& \multicolumn{3}{c}{Multi-World Performance ($K$=6)}       \\
\cmidrule(lr){2-4}
 &   AvgMinFDE & AvgMinADE & ActorMR  \\
\midrule
     1     & 1.547    &0.688  & 0.186  
    \\
       \rowcolor{customblue}
   \textbf{2}   &    \textbf{1.534}     &  \textbf{0.684}    &  \textbf{0.183}  
         \\
                     4      & 1.541  &0.687 & 0.186 
         \\
         \bottomrule
\end{tabular}
\vspace{-2mm}
\end{table}

\subsection{Ablation Studies} 
\textbf{Effects of Components.} 
We conduct ablation studies to evaluate the contribution of each component. 
As shown in Table~\ref{ablation_Components}, the baseline model without any modules demonstrates the worst performance. 
Adding Context-aware Regressor Alignment (CRA) and Spatial Cues Reconstruction (SCR) provides moderate gains by enhancing spatial relationship and disentanglement modeling. 
Replacing SCR with Motion Signals Recognition (MSR) further boosts performance by capturing motion patterns with disentanglement contextual guidance. 
Combining SCR and MSR indicates that spatial constraints are beneficial but less effective than contextual understanding. 
The full model achieves the best performance, improving AvgMinFDE, AvgMinADE, and ActorMR by 8.03\%, 9.64\%, and 12.44\% over baseline, respectively. This confirms that spatial reconstruction offers complementary geometric constraints that strengthen the synergy between contextual awareness and motion patterns.

\begin{figure*}[!tb]
    \centering
\includegraphics[width=0.93\textwidth]{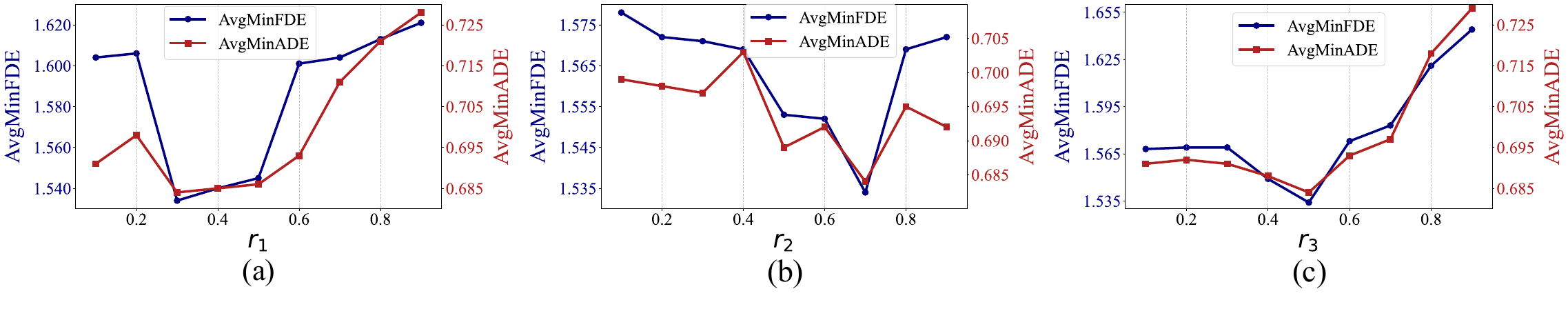}
    \caption{Ablation study on historical mask ratio (a), future mask ratio (b), and lane mask ratio (c).}
    \vspace{-0.5em}
    \label{fig:ratio}
\end{figure*}

\textbf{Effects of Mask Ratio.} 
Fig.~\ref{fig:ratio} shows the sensitivity analysis of the mask ratio. A mask ratio that is too low leads DECAMP to rely on simple interpolation during pre-training reconstruction, resulting in poor performance in downstream prediction tasks. 
Conversely, when the mask ratio is too high, excessive input removal hinders the ability of DECAMP to learn effective driving priors and lane structures, thereby degrading downstream performance. 
Our experiments show that optimal performance is achieved when 30\% of history information, 70\% of future information, and 50\% of map information are masked. 


\textbf{Effects of Encoder Depth.} 
Table~\ref{ablation_encoder} presents the results for different encoder depths. 
When comparing a depth of 4 with 2, AvgMinFDE and AvgMinADE decrease by 1.16\% and 0.73\%, respectively. 
In comparison with a depth of 6, AvgMinFDE decreases by 0.26\%. In our experiments, an encoder depth of 4 is chosen to balance model capacity and generalization. 

\textbf{Effects of Regressor Depth.} 
Table~\ref{ablation_regressor} compares model performance across different regressor depths. Increasing the depth from 1 to 2 leads to a reduction of 0.84\% in AvgMinFDE and 0.58\% in AvgMinADE. However, further increasing the depth to 4 leads to a degradation, with both metrics rising by 0.45\% and 0.44\%. 
These indicate that a depth of 2 yields the optimal performance, underscoring the importance of selecting an appropriate regressor depth. 

\textbf{Importance of Disentanglement. } 
We investigate the importance of disentanglement module. 
As shown in Table~\ref{Disentanglement}, AvgMinFDE and AvgMinADE gradually decrease as $\alpha$ increases to 2.0, reflecting modest improvements of 0.77\% and 1.29\% in displacement prediction. 
These results suggest that the model exhibits robustness against variations in $\alpha$, with $\alpha$ = 2.0 achieving the best overall performance. 

\begin{table}[!tb]
\centering
\caption{Performance sensitivity to the weighting parameter $\alpha$.}
\label{Disentanglement}
\begin{tabular}{cccccccccc}
\toprule
\multirow{2}{*}[-0.5ex]{\makecell{$\alpha$}}
& \multicolumn{3}{c}{Multi-World Performance ($K$=6)}       \\
\cmidrule(lr){2-4}
 &   AvgMinFDE & AvgMinADE & ActorMR  \\
\midrule
     0.5     &  1.546   & 0.693 & 0.189 
    \\
    1.0    & 1.539  &0.686  & 0.187 
    \\
       \rowcolor{customblue}
   \textbf{2.0}   &    \textbf{1.534}     &  \textbf{0.684}    &  \textbf{0.183}  
         \\
   3.0      & \textbf{1.534}  & 0.688& 0.185 
         \\
         \bottomrule
\end{tabular}
\vspace{-2mm}
\end{table}


\textbf{Effects of Decoder Depth.} 
Table~\ref{ablation_decoder} reports model performance across different decoder depths. 
The configuration with a 4-layer spatial decoder and a 2-layer motion decoder outperforms those with either 2-layer or 4-layer decoders, leading to reductions of 1.10\% and 0.58\% in AvgMinFDE, and 1.30\% in AvgMinADE, respectively. This aligns with the intuition that reconstructing spatial cues is more complex than modeling motion, and therefore requires a deeper decoder. 
\begin{table}[!t]
\centering
\caption{Ablation results on different depths of spatial decoder and motion decoder.}
\label{ablation_decoder}
\begin{tabular}{cccccccccc}
\toprule
\multirow{2}{*}[-0.5ex]{\makecell{Spatial \\ Decoder}}
& \multirow{2}{*}[-0.5ex]{\makecell{Motion \\ Decoder}} & \multicolumn{3}{c}{Multi-World Performance ($K$=6)}       \\
\cmidrule(lr){3-5}
 &        &        AvgMinFDE & AvgMinADE & ActorMR  \\
\midrule
   2      &     2     &  1.551   & 0.693 &   0.187
    \\
    4   &      4   &   1.543       &  \textbf{0.683}    &  0.185
         \\
    \rowcolor{customblue}
     \textbf{4}  &     \textbf{2}      &  \textbf{1.534}  & 0.684 &  \textbf{0.183} 
         \\
         \bottomrule
\end{tabular}
\end{table} 
\begin{figure}[!tb]
    \centering
\includegraphics[width=0.91\linewidth]{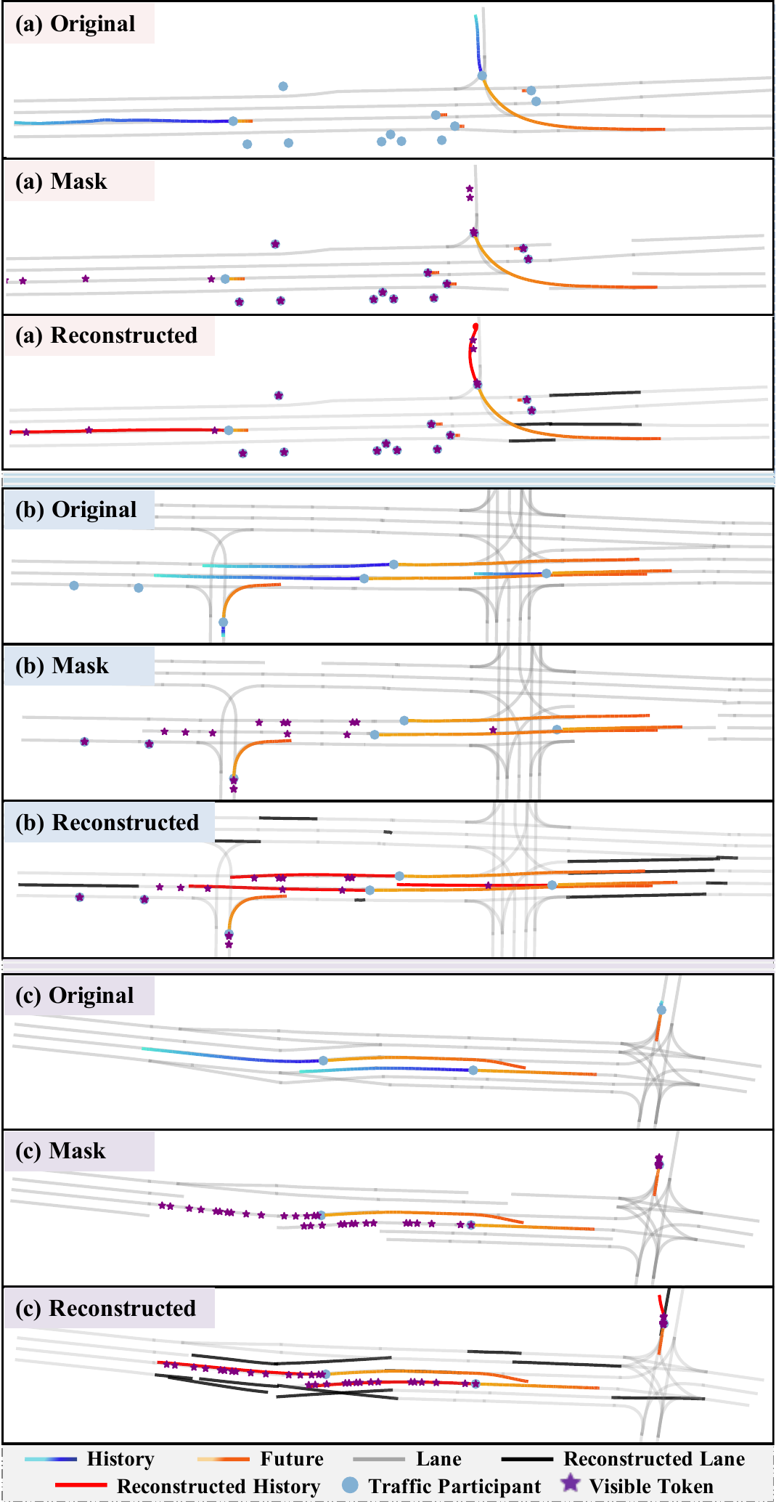}
    \caption{Representative reconstruction results on Argoverse 2 dataset, with historical states and lane segments masked for clarity.}
     \vspace{-0.5em}
    \label{fig:recon2}
\end{figure} 

\subsection{Qualitative Analysis} 
\textbf{Reconstruction.} 
Fig.~\ref{fig:recon2} presents reconstruction results across three independent scenes, which are labeled as (a)-(c). 
For each scene, we display three views: 
Original (full scene), Mask (a subset of historical trajectories and lane segments are masked to simulate partial observability), and Reconstructed (recovery by DECAMP from the masked input). 
Across scenes, the reconstruction effectively restores the missing elements and preserves key spatial structures, demonstrating that our method has learned meaningful behavioral priors and lane topology during pre-training. 


\textbf{Multi-World Prediction.} 
\begin{figure}[!tb]
    \centering
\includegraphics[width=0.91\linewidth]{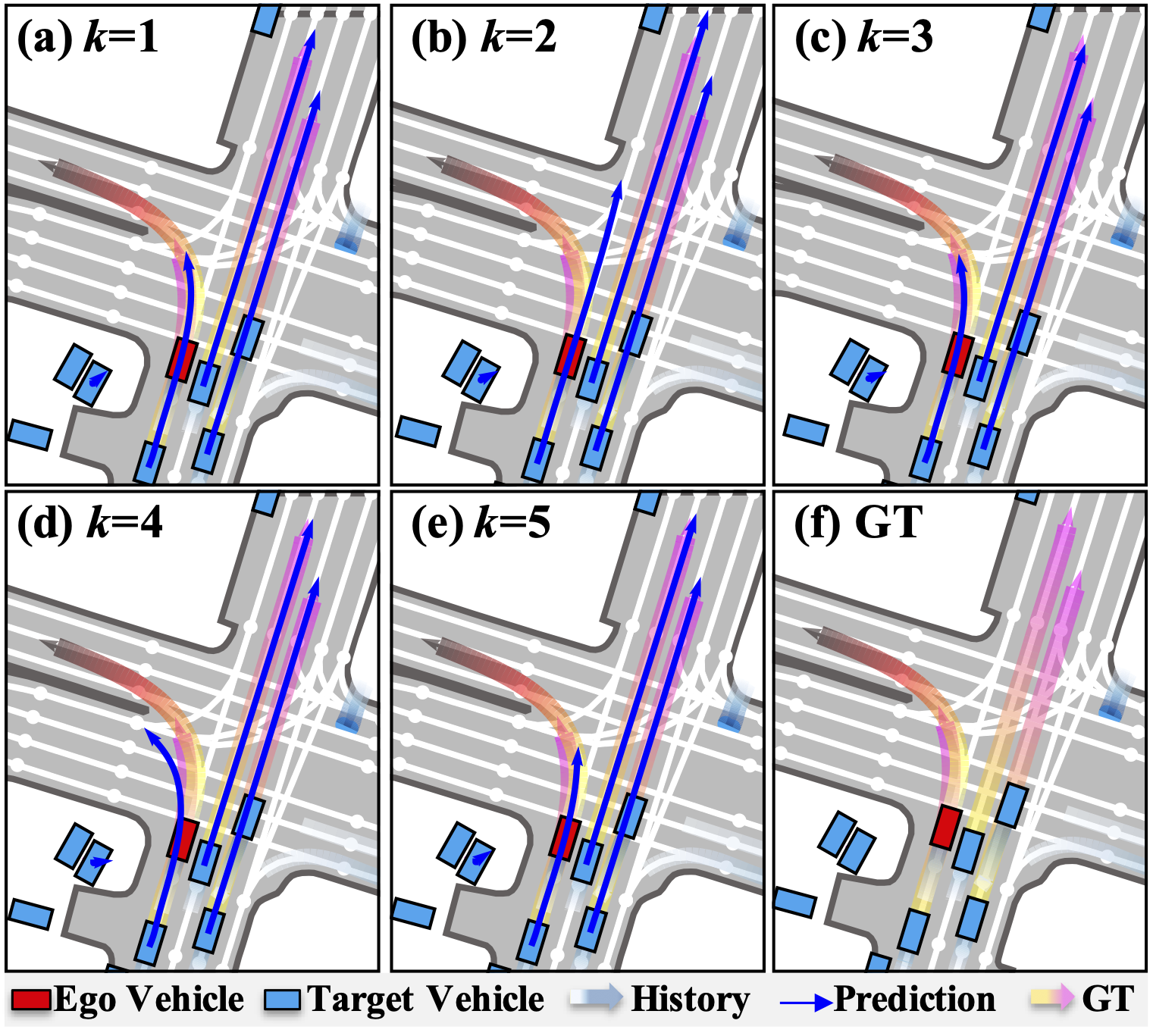}     \caption{Visualization of top-$K$ scene-level predictions ($K$=5), where $k=\{1,2,\dots,5\}$ denotes a distinct predicted scene-level modality, and GT denotes the ground truth.}
    \label{prediction1.2}
\end{figure} 
Fig.~\ref{prediction1.2} illustrates scene-level predictions for multiple agents. The visualization indicates that our model leverages prior knowledge from pre-training to generate scene-consistent joint predictions, closely aligning with ground truth direction and velocity. 
In the second prediction, the model further captures diverse driving intentions, demonstrating its ability to represent multiple potential future trajectories. 

\section{Conclusion} 
This paper introduces DECAMP, a context-aware pre-training framework for multi-agent motion prediction. 
DECAMP disentangles behavioral pattern learning from latent feature reconstruction, enabling interpretable and transferable representations that closely align with real-world driving behaviors. 
Moreover, integrating collaborative spatial–motion pretext tasks enhances the ability of model to capture both spatial context and dynamic intention, yielding richer behavioral priors. 
Experimental results demonstrate that DECAMP achieves competitive performance compared with existing multi-world methods, highlighting its effectiveness for joint prediction.

\bibliographystyle{IEEEtran}
\bibliography{IEEEabrv}

\end{document}